\documentclass[letterpaper, 10 pt, conference]{ieeeconf}  

\usepackage{amssymb}
\usepackage{amsmath}
\usepackage{algorithmic}
\usepackage{algorithm}
\usepackage{array}
\usepackage{amsfonts}
\usepackage{color}
\usepackage{caption}
\usepackage{bm} 
\usepackage{booktabs}
\usepackage{multirow}
\usepackage{makecell}
\usepackage[table]{xcolor}
\usepackage{textcomp}
\usepackage{stfloats}
\usepackage{url}
\usepackage{verbatim}
\usepackage{graphicx}
\usepackage{wrapfig}
\usepackage{floatflt}
\usepackage{url}
\usepackage{breakurl}
\usepackage[colorlinks=true, linkcolor=red, urlcolor=blue, citecolor=green]{hyperref}

\IEEEoverridecommandlockouts                              

\title{\LARGE \bf
G$^{3}$CN: Gaussian Topology Refinement Gated Graph Convolutional Network for Skeleton-Based Action Recognition
}

\author{Haiqing Ren$^{1}$, Zhongkai Luo$^{1}$, Heng Fan$^{2}$, Xiaohui Yuan$^{2}$, Guanchen Wang$^{3}$ and Libo Zhang$^{1,*}$
\thanks{$^{1}$ Institute of Software, Chinese Academy of Sciences, Beijing, 100190, China}
\thanks{$^{2}$ Department of Computer Science and Engineering, University of North Texas, Denton, TX 76206, USA}
\thanks{$^{3}$ Chadwick School 26800 South Academy Drive, Palos Verdes Peninsula, CA 90274, USA}
\thanks{$^{*}$ Corresponding author (libo@iscas.ac.cn). Libo Zhang was supported by National Natural Science Foundation of China (No. 62476266) and the Key Research Program of Frontier Sciences, CAS, (No. ZDBS-LY-JSC038). Heng Fan and Xiaohui Yuan are not supported by any funds for this work.}}

\providecommand{\mykeywords}[1]{\textbf{\textit{Keywords---}} #1}

\begin{document}

\maketitle
\thispagestyle{empty}
\pagestyle{empty}

\begin{figure*}[b]
    \centering
    \includegraphics[width=0.75\linewidth]{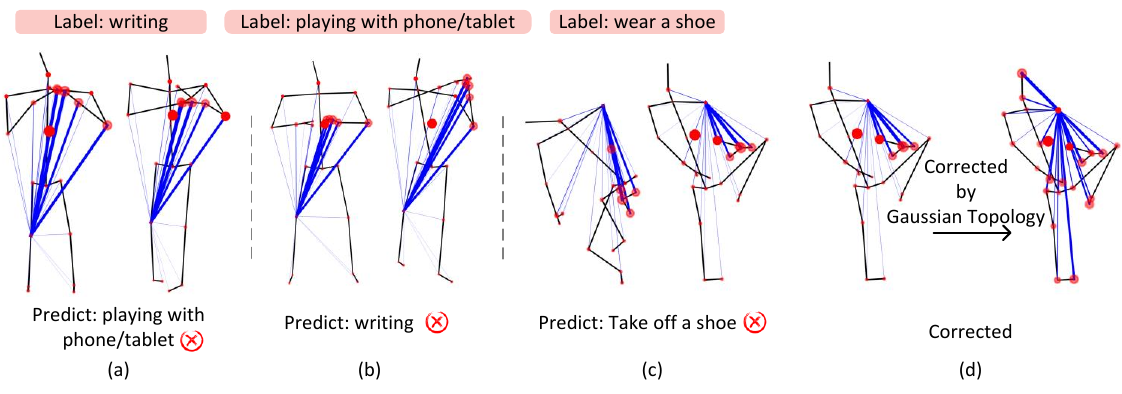}
    \vspace{-1.0em}
    \caption{Given a skeleton sequence, the joint connections form a topology graph. In this figure, we visualize topology graphs learned by CTR-GCN \cite{Chenctrgcn}. For "playing with phone/tablet" and "typing on a keyboard," only correlations between the "left knee" and other joints are shown, while for "wear a shoe," correlations with the "left shoulder" are visualized. The size of the red circles and width of the blue lines represent correlation strength. In CTR-GCN’s graphs for "wear a shoe," the correlation between the "left shoulder" and foot joints is weak (shown in (c)), which is corrected by the Gaussian topology method (shown in (d)).}
    \label{fig:enter-label-motivation}
    \vspace{-5mm}
\end{figure*}
\begin{abstract}

Graph Convolutional Networks (GCNs) have proven to be highly effective for skeleton-based action recognition, primarily due to their ability to leverage graph topology for feature aggregation, a key factor in extracting meaningful representations. However, despite their success, GCNs often struggle to effectively distinguish between ambiguous actions, revealing limitations in the representation of learned topological and spatial features. To address this challenge, we  propose a novel approach, Gaussian Topology Refinement Gated Graph Convolution (G$^{3}$CN), to address the challenge of distinguishing ambiguous actions in skeleton-based action recognition. G$^{3}$CN incorporates a Gaussian filter to refine the skeleton topology graph, improving the representation of ambiguous actions. Additionally, Gated Recurrent Units (GRUs) are integrated into the GCN framework to enhance information propagation between skeleton points. Our method shows strong generalization across various GCN backbones. Extensive experiments on NTU RGB+D, NTU RGB+D 120, and NW-UCLA benchmarks demonstrate that G$^{3}$CN effectively improves action recognition, particularly for ambiguous samples. The code is available at \url{https://github.com/CyanSea123/G3CN-Gaussian-Topology}.

\end{abstract}

\mykeywords{\textbf{Skeleton-based action recognition, Gaussian topology, Gated graph convolution, G$^{3}$CN.}} 

\section{INTRODUCTION}

Human action recognition is crucial for applications like human-robot interaction and video surveillance, enhancing user experience and efficiency. Skeleton-based recognition has gained popularity due to advances in deep sensors and its robustness in complex environments \cite{MST-GCN, lstm, hh-to,  gcl-sk1}. Among various methods, GCN is widely used for modeling human skeleton topology \cite{Liucv20,  Ddgcn, Zhaocv36, Tangcv27}. As GCN-based methods evolve, key approaches for refining topology graphs include 2s-AGCN \cite{2SShicv24}, InfoGCN \cite{Info-GCN}, and CTR-GCN \cite{Chenctrgcn}.

Despite advancements in modeling skeleton topology graphs (correlations between joints) , distinguishing ambiguous actions like "playing with a phone/tablet," "writing," "wear a shoe," and "take off a shoe" remains challenging. These actions involve similar joint movements, especially for actions like wearing and removing shoes. Following the previous method CTR-GCN\cite{Chenctrgcn}, we visualize adaptive graphs for "writing" and "playing with a phone/tablet" in Fig.\ref{fig:enter-label-motivation}. For both actions, hand movements are key for differentiation, but the correlations between joints of one hand and others are strong, while those for the other hand are weak. This limits the model’s ability to differentiate the actions. Similarly, for "wear a shoe," the weak correlation between foot joints and other joints hinders accurate identification, as shown in Fig.\ref{fig:enter-label-motivation}(c).

The weak correlation is mainly due to individual differences in performing actions and sensor noise. For example, in the writing action, one hand's movement may be captured clearly, while errors in tracking the other hand introduce noise. This distorts the topology graph, weakening the expected correlation between key joints.


\begin{figure}[t]
    \centering
    \includegraphics[width=1\linewidth]{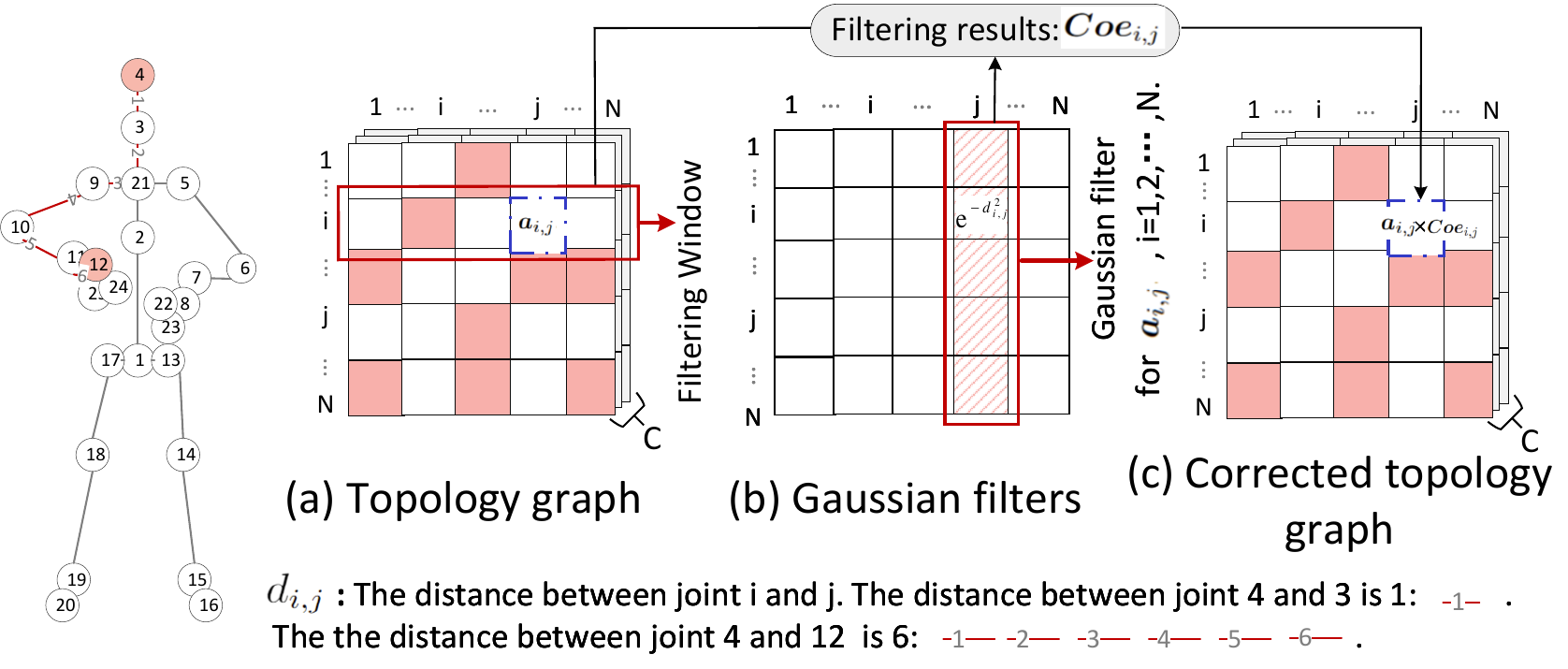}
    \caption{Filtering the topological graph with a Gaussian filter. }
    \label{fig:enter-label-gaussain}
    \vspace{-1.5em}
\end{figure}
In this paper, we introduce a method called Gaussian topology, inspired by Gaussian filtering, to improve topology graph modeling. We treat each topology graph as an image, where each pixel represents the correlation between two joints, as shown in Fig.\ref{fig:enter-label-gaussain} (a). Noise can cause high-correlation pixels to appear weak, so we apply Gaussian filtering to adjust the value of the current pixel based on its neighboring values. Specifically, we correct the correlation between two joints, i.e., correlation from joint $i$ to joint $j$, by considering the correlations between joint $i$ and its neighboring joints. 

\textbf{Gaussian filters.} As shown in Fig.\ref{fig:enter-label-gaussain} (b), `Gaussian filters' is a matrix $\Phi\in{\mathbb{R}}^{{N}\times{N}}$, where $N$ denotes the total number of joints. In matrix $\Phi$, the element in the $i$th row and $j$th column is $e^{-d_{i,j}^2}$, where $d_{i,j}$ represents the distance between joint $i$ and joint $j$, as shown in Fig.\ref{fig:enter-label-gaussain}.

\textbf{Topology graph.} For a given action, the correlations between joints (topology graph) is formulated as a matrix $\bm{A}\in{\mathbb{R}}^{{C}\times{N}\times{N}}$, where $C$ denotes the channels of topology graph. The element $\bm{a}_{i,j (i=1,...,N; j=1,...,N)}\in\bm{A}$ represents the correlation from joint  $i$ to joint $j$, as shown in Fig.\ref{fig:enter-label-gaussain}(a).

The $j$th column vector in matrix $\Phi$ is the filter kernel for $\bm{a}_{i,j (i=1,...,N)}$ and the size of filter widow is $1\times{N}$, as shown in Fig.\ref{fig:enter-label-gaussain}.

\textbf{Correction coefficient.} Given two joints $i$ and $j$, we define the result of Gaussian filtering at the $(i, j)$ position in the topological graph as the correction coefficient for the correlation between these two joints, denoted as $\bm{Coe}_{i,j}$. By using the correction coefficient $\bm{Coe}_{i,j}$, the correlation between these two joints are corrected, as shown in Fig.\ref{fig:enter-label-gaussain}(c).

Besides, in traditional GCN-based methods, the current joint’s features are aggregated by summing those of correlated joints. We propose using GRU \cite{Cho} for feature aggregation, which filters out redundant information, resulting in sparser and more relevant features that enhance the model's representation capability. 

In summary, the contributions of this paper are as follows:

\begin{itemize}
\item [(1)] We propose a Gaussian topology refinement gated graph convolution (G$^{3}$CN) that models topology dynamically using Gaussian filters and aggregates features sparsely with gates, improving action recognition performance for ambiguous actions. 
\item [(2)] The proposed G$^{3}$CN has the universality to be compatible with most GCN based backbones.
\item [(3)] Extensive experiments on NTU RGB+D, NTU RGB+D 120, and NW-UCLA datasets show that G$^{3}$CN enhances the representation ability of GCN models for ambiguous actions. 
\end{itemize}

\section{RELATED WORK}

\subsection{GCN-based Skeleton Action Recognition}

GCN-based methods for Skeleton Action Recognition use GCNs for spatial feature extraction and TCNs for temporal features \cite{Yancv32}. The ST-GCN framework by \cite{Yancv32} introduced graph-based formulations to model dynamic skeletons, utilizing the natural graph structure of human joints and bones \cite{Chenctrgcn, Liucv20, Yancv32, Zhangpr42}. GCN-based methods can be broadly categorized into two types: (1) Topology modeling and (2) Other techniques.

\textbf{Topology modeling research:}
Topology modeling aims to capture joint correlations for aggregating features in GCNs. Early methods used static topologies based on natural joint and bone connections, as seen in ST-GCN by Yan et al. \cite{Yancv32}. However, static topologies struggle with capturing dynamic semantic relationships in actions, such as hand-head or hand-hand correlations in actions like drinking water or clapping.

To address this, dynamic topologies have been proposed. For example, Li et al. \cite{Licv15} introduced Alinks, while Shi et al. \cite{2SShicv24} and Zhang et al. \cite{Zhangiccv35} used self-attention mechanisms. Ye et al. \cite{Dynamic-GCN} integrated contextual features from all joints to improve correlation learning. Cheng et al. \cite{ChengADG} proposed DCGCN, which tailored topologies for different channel groups, and Chen et al. \cite{Chenctrgcn} introduced CTR-GCN for effective channel-wise topology refinement. Pang et al. \cite{PangIGformer} enhanced topology modeling for multi-person action recognition.

\textbf{Others:}
Some studies focus on machine learning methods, such as contrastive learning, to improve recognition accuracy \cite{gcl-contrastive,ZhouFrhead}. Zhou et al. \cite{ZhouFrhead} proposed a feature refinement module based on contrastive learning, while Huang et al. \cite{gcl} guided graph learning with cross-sequence context for better intra-class compactness and inter-class separation.

However, distinguishing ambiguous actions remains challenging. To address this, we propose a novel Gaussian topology method, which uses a Gaussian filter to correct joint correlations by considering neighboring joint interactions, thereby improving topology representation. This is the first attempt to incorporate Gaussian filtering for skeleton action recognition.

\subsection{Graph Convolutional Networks}
Convolutional neural networks excel at processing Euclidean data like images, but there is growing interest in graph convolutional networks (GCNs) for non-Euclidean data. GCNs are divided into spectral and spatial methods. Spectral methods use spectral analysis for locality \cite{Henaff, Duvenaud, Kipf, Yancv32}, while spatial methods apply convolutional filters to graph nodes and neighbors. This paper focuses on the spatial method for GCNs, specifically the feature aggregation step. 

\section{METHOD}
\label{sec:blind}

In this section we will give the details of our proposed G$^{3}$CN. The frame work of it is depicted in  Fig.\ref{fig:enter-label-framework}.
\subsection{Preliminary}

\textbf{Graph Construction.} A human skeleton is modeled as a graph, with joints as vertices and bones as edges. The graph $\mathcal{G} = (\mathcal{V}, \mathcal{E}, \mathcal{X})$ represents correlations between joints, where $\mathcal{V}= ({{v}_1, {v}_2, ......,{v}_N}) $ is the set of $N$ vertices and $\mathcal{X}$ is the feature set. Each vertex $v_i$ has a feature dimension of $C$. A skeleton sequence with $T$ frames is represented as ${\bm{X}}\in{\mathbb{R}}^{{T}\times{N}\times{C}}$. The edge set $\mathcal{E}$ is represented by the adjacency matrix $\bm{A}$.

\textbf{Graph Convolution for Skeleton-Based Action Recognition.} GCN models combine graph and temporal convolutions to capture spatial and temporal information of skeleton actions. The graph $\mathcal{G}$ governs information exchange between joints through the adjacency matrix $\bm{A}$. In static methods, $\bm{A}$ is treated as trainable parameters $\bm{A}_{static}\in {\mathbb{R}}^{{N}\times{N}} $. In adaptive GCNs, such as CTR-GCN \cite{Chenctrgcn}, $\bm{A}$ is learned based on the input sample and structured as $\bm{A}\in {\mathbb{R}}^{{C^{\prime}}\times{N}\times{N}}$, with each channel having its independent topology. The graph convolution is defined as

\begin{equation}
{\bm{X}_{update}}=\bm{A}\bm{X}{\bm{W}} \label{1}
\end{equation}
where $\bm{X}\in{\mathbb{R}}^{{T}\times{N}\times{C}}$ denotes the input feature, ${\bm{X}_{update}}\in{\mathbb{R}}^{{T}\times{N}\times{C^{\prime}}}$ denotes the updated representation using graph convolution, and ${\bm{W}}\in{\mathbb{R}}^{{1}\times{1}\times{C}\times{C^{\prime}}}$ is the weight of the ${1}\times{1}$ convolution for feature transformation.

\subsection{Gaussian Topology}
Given an adjacency matrix (topology graph) $\bm{A}\in{\mathbb{R}}^{{C^{\prime\prime}}\times{N}\times{N}}$, $\bm{a}_{i,j}$$ (i=1,...,N;j=1,...,N)\in\bm{A}$ represents the correlation from joint  ${v}_i$ to  ${v}_j$ ($\bm{a}_{j,i}$  represents the correlation from joint ${v}_j$ to  ${v}_i$).  

\begin{figure*}[t]
    \centering
    \includegraphics[width=0.7\linewidth]{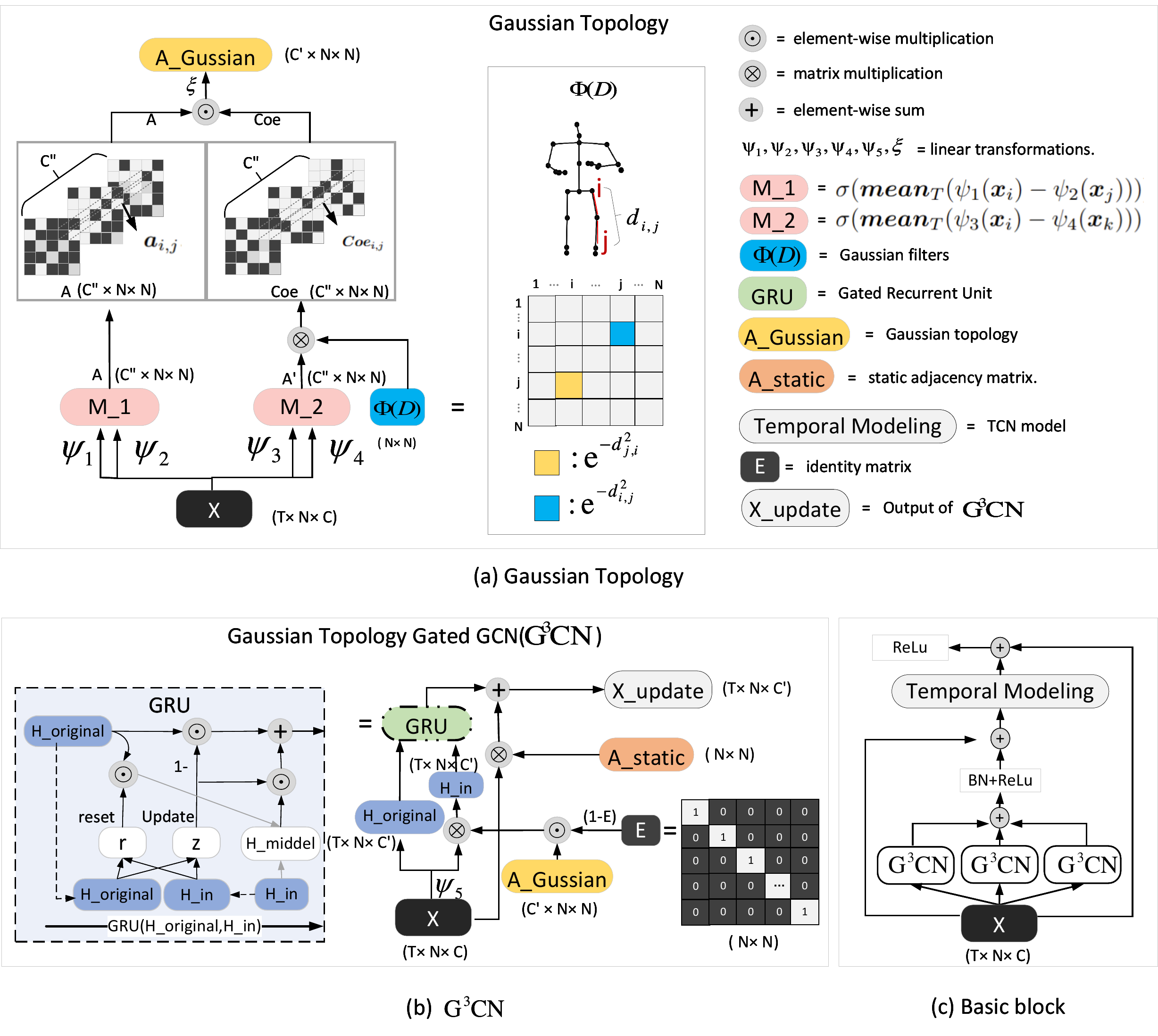}
    \caption{The framework of the proposed G$^{3}$CN. (a) Gaussian Topology. (b) The framework of the proposed G$^{3}$CN. (c) The basic block of our approach builds on the backbone of \cite{Chenctrgcn}, where GTR-GCN is changed to  G$^{3}$CN.}
    \label{fig:enter-label-framework}
    \vspace{-1.5em}
\end{figure*}

For a pair of joints ${v}_i$ and ${v}_j$, their corresponding features $\bm{x}_{i}\in {\mathbb{R}}^{{T}\times{C}}$ and $\bm{x}_{j}\in {\mathbb{R}}^{{T}\times{C}}$. We calculate distances between features of different joints along the channel dimension and utilize the nonlinear transformations of these distances as preliminary correlation from ${v}_i$ to ${v}_j$, formulated as
\begin{equation}
\bm{a}_{i,j}=\sigma(\bm{mean}_{T}(\psi_1(\bm{x}_{i})-\psi_2(\bm{x}_{j})))\label{2}
\end{equation}
where  $\psi_1$ and $\psi_2$ represent linear transformations, $\sigma$ denotes the activation function (specifically $tanh$ in this paper). The function $\bm{mean}_{T}(\cdot)$ calculates the mean value of features across all frames.  We employ linear transformations $\psi_1$ and $\psi_2$ to reduce feature dimension.

Different from the previous method, we correct the correlation $\bm{a}_{i,j}$ between two joints by using the correlations between the joint $i$ and its neighboring joints. 

First, we define another adjacency matrix (topology graph), denoted as $\bm{A}^{\prime}\in {\mathbb{R}}^{{C^{\prime\prime}}\times{N}\times{N}}$. Subsequently, Gaussian filtering will be applied to this topology graph to obtain correction coefficients for the correlations between joints in the original adjacency matrix (topology graph) $\bm{A}$. ${\bm{a}^{\prime}}_{i,j}$ in the adjacency matrix $\bm{A^{\prime}}$ is expressed as
\begin{equation}
{\bm{a}^{\prime}}_{i,j}=\sigma(\bm{mean}_{T}(\psi_3(\bm{x}_{i})-\psi_4(\bm{x}_{j})))\label{4}
\end{equation}
where $\psi_3$, $\psi_4$ denote linear transformations.We employ linear transformations $\psi_1$ and $\psi_2$ to reduce feature dimension.

Second, we define the Gaussian filters as follows.

\textbf{Gaussian filters.} We hypothesize that the Gaussian filter kernel at position $(i, j)$ of the topology graph $\bm{A^{\prime}}$ follows a normal distribution, and the size of filter widow is $1\times{N}$. For simplicity, we use the definition of a standard normal distribution with variance $1/(2\pi)$\cite{Stigler1982}, whose probability density function is $\varphi(d) = e^{\_ \pi{d^2}}$, where $d$ is the random variable, i.e., the distance between joints. Then, we define the Gaussian filters as $\bm{\Phi(D)}\in{\mathbb{R}}^{{N}\times{N}}$, where $\bm{D}\in{\mathbb{R}}^{{N}\times{N}} $, and the element  $d_{i,j}\in \bm{D}$ is the distance between  ${v}_i$ and  ${v}_j$. The distance between two joints is defined as shortest path between two joints as shown in Fig.\ref{fig:enter-label-gaussain}. The element $\varphi_{i,j}\in \bm{\Phi(D)}$ is formulated as
\begin{equation}
\varphi_{i,j}= e^{-{d_{i,j}^2}} \label{3}
\end{equation}

The $j$th column of Gaussian filters $\bm{\Phi(D)}$ denoted as $\bm{\varphi}_{j}$ is the Gaussian filter kernel at position $(i, j)$ of the topology graph $\bm{A^{\prime}}$. $\bm{\varphi}_{j}$$=(\varphi_{1,j}, \varphi_{2,j}$ $,...,$ $ \varphi_{j,j},..., \varphi_{N,j})$. Following the Eq.\ref{3}, the weights ($\varphi_{i,j (i=1,2,...,N)} $) of the Gaussian filter kernel $\bm{\varphi}_{j}$ decreases as the distance between joint $j$ and other joints increases. 

Third, we formulate the correction coefficient for the current correlation using the Gaussian filtering.

\textbf{Correction coefficient.} For the topology graph $\bm{A^{\prime}}$, we define the result of Gaussian filtering at the $(i, j)$ position as the correction coefficient for the correlation from joint  ${v}_i$ to  ${v}_j$, denoted as $\bm{Coe}_{i,j}$. $\bm{Coe}_{i,j}$ is formulated as

\begin{align}
\bm{Coe}_{i,j}&=\bm{\varphi}_{j} \ast \bm{a}^{\prime}_{i}\nonumber  \\
&=\sum_{k=1}^{N}{\varphi_{k,j}} \odot {\bm{a}^{\prime}}_{i,j} \label{5}
\end{align}
where $\bm{\varphi}_{j}$ denotes the $j$th column of Gaussian filters $\bm{\Phi(D)}$, $\bm{a}^{\prime}_{i}$ denotes the $i$th column of topology graph (adjacency matrix) $\bm{A}^{\prime}$, $\bm{\varphi}_{j} \ast \bm{a}^{\prime}_{i}$ presents Gaussian filtering at the $(i, j)$ position of the topological graph  $\bm{A}^{\prime}$. Finally, the correction coefficient is normalized as 
\begin{equation}
\bm{\bar{Coe}}_{i,j}=\frac{\bm{Coe}_{i,j}}{\mathop{max}\limits_{j\in (1,...,N)}\left|\bm{Coe}_{i,j}\right| }\label{6}
\end{equation}

Lastly, we obtain the Gaussian topology by multiplying the correction coefficients and the original topology.

\textbf{Gaussian Topology.}  After considering the correction coefficient, the correlation $\bm{\bar{a}}_{i,j}$  is formulated as
\begin{equation}
\bar{\bm{a}}_{i,j}=\xi(\bm{a}_{i,j}\odot\bm{\bar{Coe}}_{i,j})\label{7}
\end{equation}
where $\odot$ denotes element-wise multiplication, $\xi$ denotes linear transformations. Finally, the Gaussian topology is defined as $\bm{A}_{Gaussian}\in {\mathbb{R}}^{{C^{\prime}}\times{N}\times{N}} $,  where each element at position $(i, j)$ is  $\bm{\bar{a}}_{i,j}$.

Following Eq.\ref{1}, the updated representation of current joint is formulated as
\begin{equation}
{\bm{x}}_{i}^{update}=\sum_{j=1}^{N}\bm{\bar{a}}_{i,j} \odot \bm{x}_j{\bm{W}} \label{8}
\end{equation}

\subsection{Gated GCN Block}
In Eq.\ref{8}, the updated representation $\bm{x}_{i}^{update}$ of joint ${v}_i$ is obtained by summing the features of correlated joints with the corresponding weight $\bm{a}_{i,j}$, akin to message passing. However, this may lead to redundancy, as the current joint receives all messages without selectivity.

To improve this, we use a GRU (Gated Recurrent Unit), which performs message passing through three operations: message transmission (M), update (U), and read (R). Here, the feature of the current joint is treated as the original state, and the features of related joints are the inputs. Thus, ${\bm{X}}$ represents the original state of all joints, denoted as ${\bm{H}}_{original}$, while the inputs are ${\bm{H}}_{in}$. The input features are updated according to the GRU formulation.

\begin{subequations}
\begin{align}
&{\bm{H}}_{original}=\bm{X} \label{Za}\\
&{\bm{H}}_{in}=((1-\bm{E})\odot\bm{A}_{Gaussian})\bm{X}\bm{W}\label{Zb}\\
&\bm{Z}=\sigma({\bm{H}}_{original}\bm{W}_{(zo)}+{\bm{H}}_{in}\bm{W}_{(zi)})\label{Zc}\\
&\bm{R}=\sigma({\bm{H}}_{original}\bm{W}_{(ro)}+{\bm{H}}_{in}\bm{W}_{(ri)})\label{Zd}\\
&{\bm{H}}_{middle}=tanh(\bm{R}\odot({\bm{H}}_{in}\bm{W}_{(mi)})+{\bm{H}}_{original}\bm{W}_{(mo)} )\label{Ze}\\
&{\bm{H}}_{final}=(1-\bm{Z})\odot{\bm{H}}_{middle}+\bm{Z}\odot{\bm{H}}_{original}\label{Zf}\\
&{\bm{X}}_{update}={\bm{H}}_{final}+\bm{A}_{static}\bm{X}{\bm{W}}\label{Zg}
\end{align}
\end{subequations}
where $\bm{W}_{(zo)}$,  $\bm{W}_{(zi)}$, $\bm{W}_{(ro)}$,  $\bm{W}_{(ri)}$, $\bm{W}_{(mo)}$,  $\bm{W}_{(mi)}$ denote weight matrices in GRU, $\bm{E}\in{\mathbb{R}}^{{N}\times{N}}$ represents the identity matrix. 

Compared to the original GCN in Eq. \ref{1}, the gated GRU filters out redundant information through its gates, resulting in more sparse aggregated features.

\subsection{Model Architecture}
The proposed model takes as input a sequence of skeletons represented by ${\bm{X}_{in}} \in \mathbb{R}^{T \times N \times 3}$, where $T$ denotes the number of frames, $N$ represents the number of joints, and each joint has three-dimensional coordinates ${x, y, z}$. Our approach builds upon the framework introduced in \cite{Chenctrgcn}, with the modification of replacing GCN modules (CTR-GCNs) with our proposed G$^{3}$CN, illustrated in Fig.\ref{fig:enter-label-framework}(c). In our model architecture, temporal CNNs (TCNs) are used to extract temporal features, while G$^{3}$CNs focus on extracting spatial features across the spatial dimensions. The model consists of 10 basic blocks, followed by a pooling layer to derive high-level feature vectors. Subsequently, a classification head with a softmax activation function generates predictions $\vec{\bm{y}} \in \mathbb{R}^{C_k}$, where ${C_k}$ denotes the number of classes. The training objective is governed by the cross-entropy loss $\mathcal{L}$
\begin{equation}
\mathcal{L}=-\sum_{i}y_{i}log\vec{\bm{y}}_{i} \label{10}
\end{equation}
where $\bm{y}$ denotes the the ground truth label. It is important to note that the implementation details of the backbone are not the primary focus of our proposed method. The experimental section will demonstrate that the proposed  G$^{3}$CN can improve most GCN based models.

\section{EXPERIMENTS}
\subsection{Datasets}

\begin{table*}[t]
  \centering
  \caption{Top-1 accuracy (\%) of different skeleton-based action recognition methods on NTU 60 and NTU120
datasets. 
}
    \footnotesize
    \scalebox{0.75}{
 \begin{tabular}{c||c|c|c|c|c|c|c|c||c|c|c|c|c|c|c|c}
    \toprule
    \multicolumn{1}{c||}{\multirow{3}{*}{\textbf{Method}}} & \multicolumn{8}{c||}{\textbf{NTU RGB+D}} & 
     \multicolumn{8}{c}{\textbf{NTU RGB+D 120}} \\     
     \cmidrule{2-17}   
     \multicolumn{1}{c||}{} & \multicolumn{4}{c|}{X-Sub} & \multicolumn{4}{c||} 
      {X-View} & \multicolumn{4}{c|}{X-Sub} & \multicolumn{4}{c}{X-Set}  \\
      \cmidrule{2-17}   
     \multicolumn{1}{c||}{} & \multicolumn{1}{c|}{J} & \multicolumn{1}{c|} {B} & \multicolumn{1}{c|}{J+B} & \multicolumn{1}{c|}{4S} & \multicolumn{1}{c|}{J} & \multicolumn{1}{c|} {B} & \multicolumn{1}{c|}{J+B} & \multicolumn{1}{c||}{4S} & \multicolumn{1}{c|}{J} & \multicolumn{1}{c|} {B} & \multicolumn{1}{c|}{J+B} & \multicolumn{1}{c|}{4S} & \multicolumn{1}{c|}{J} & \multicolumn{1}{c|}{B} & \multicolumn{1}{c|}{J+B} & \multicolumn{1}{c}{4S} \\
    \midrule
     Shift-GCN\cite{ShiftGCN} &87.8& -  &89.7 &90.7& 95.1 &  -  &96.0& 96.5& 80.9 &   -   & 85.3 &85.9& 83.2 &  -    &    86.6&87.6 \\
   Dynamic GCN\cite{Dynamic-GCN} &-&  - &   -  & 91.5   &  -    &  -  &- &    96.0 & -     &    - &   -   &   87.3 &   -   &   -   &   -   &    88.6 \\
     MST-GCN\cite{MST-GCN}& 89.0& 89.5& 91.1& 91.5& 95.1 &95.2& 96.4& 96.6& 82.8 &84.8 &87.0 &87.5& 84.5 &86.3 &88.3 &88.8 \\
     STC-net\cite{STCnet_2023_ICCV}& -  &  -    &   92.5   & 93.0     & -     &   -   &  96.7    &97.1      &  -  & - &   89.3   & 89.9     & -    &- &    90.7  & 91.3 \\
     HD-GCN\cite{HDGCN_2023_ICCV}& -  &   -   & 92.4     &93.0     &  -    &  -    & 96.6     & 97.0     &    - & - & 89.1  &  89.8      & - & -   &     90.6 &91.2   \\
     L-kernelGCN\cite{Large-kernel}& 89.0 & -  & 90.7   &  -    &  95.0&-    & 96.1    & -     &82.6      &-    & 86.3&  - &84.0& -& 87.8 &-   \\
    FR-Head\cite{ZhouFrhead} &90.3&   91.0&     &92.8     &   95.2   &   95.0   & -   &  96.8   &  85.5  &   86.8  &  -  &89.5       &      87.3 &    88.1  &   -  &90.9   \\
     SkeletonGCL\cite{gcl}&90.8   & 91.1     & 92.6     &  93.1    &  95.5&95.7    &  96.6    & 97.1     &85.6      &  87.4    & 89.2 &  89.8 &87.3& 88.7& 90.5 &91.2   \\
     
     DeGCN\cite{DeGCN}& - & -  & -   &  93.6    &  -&-    & -    & 97.4     &-      &-    & -&  91.0 &-& -& - &92.1   \\
     SelfGCN\cite{Self-GCN}& - & -  & -   &  93.1    &  -&-    & -    & 96.6     &-      &-    & -&  89.4 &-& -& - &91.0   \\

      \midrule     
 2s-AGCN\cite{2SShicv24} & 88.9& 89.2 &91.0 &91.5 &94.5 &94.1& 95.7& 95.9& 84.0 &85.1 &87.8& 88.2& 85.3& 86.3 &89.0 &89.6  \\
 \rowcolor{gray!40} \makecell {2s-AGCN\cite{2SShicv24} \\ w/G$^{3}$CN}&  \makecell{\textcolor{red}{90.1}\\$\uparrow$\textbf{ \textcolor{red}{1.2}}   }&  \makecell{\textcolor{red}{90.2}\\$\uparrow$\textbf{ \textcolor{red}{1.0}} }   &  \makecell{91.8\\$\uparrow$0.8}    &    \makecell{92.3\\$\uparrow$0.8}  & \makecell{94.9\\$\uparrow$0.5}     &   \makecell{94.8\\$\uparrow$0.7}   &   \makecell{96.1\\$\uparrow$ 0.4}  &\makecell{96.4\\$\uparrow$ 0.5}  &  \makecell{84.6\\$\uparrow$ 0.6}     & \makecell{85.7\\$\uparrow$ 0.6}     & \makecell{88.0\\$\uparrow$0.2} &  \makecell{88.4\\$\uparrow$0.2}    &    \makecell{86.1\\$\uparrow$ 0.8}   &\makecell{86.9\\$\uparrow$ 0.6}       &  \makecell{89.3\\$\uparrow$ 0.3}    & \makecell{89.7\\$\uparrow$ 0.1}  \\
    \midrule
    CTR-GCN\cite{Chenctrgcn} &89.7&   90.3&  92.0  &  92.4  &    94.8   &   94.8   &    96.3 &  96.8    &    84.7  &     85.7  &   88.7 &      88.9 &      86.7 &     87.5  &    90.1  &   90.5\\
    \rowcolor{gray!40} \makecell { CTR-GCN\cite{Chenctrgcn} \\ w/G$^{3}$CN}& \makecell{\textcolor{red}{90.8}\\$\uparrow$ \textbf{ \textcolor{red}{1.1}} }    & \makecell{90.9\\$\uparrow$ 0.6}     &   \makecell{92.4\\$\uparrow$ 0.4}   &       \makecell{92.9\\$\uparrow$ 0.5} & \makecell {95.2\\$\uparrow$ 0.4}     &  \makecell{95.1\\$\uparrow$ 0.3}     &   \makecell{96.3}    &    \makecell{96.7}   & \makecell {85.1\\$\uparrow$ 0.4}    & \makecell{86.6\\$\uparrow$ \textbf{0.9}} &  \makecell{88.7}    &\makecell{89.1\\$\uparrow$ 0.2}    &\makecell{ 87.1 \\$\uparrow$ 0.4}&  \makecell{ 88.0\\$\uparrow$ 0.5}&  \makecell{90.5 \\$\uparrow$ 0.4}  &   \makecell{ 90.9 \\$\uparrow$ 0.4}     \\
     \midrule
  InfoGCN\cite{Info-GCN} &89.8& 90.6& 91.6& 92.7 &95.2& 95.5& 96.5 &96.9& 85.1& 87.3 &88.5 &89.4& 86.3 &88.5& 89.7& 90.7\\
    \rowcolor{gray!40} \makecell { InfoGCN\cite{Info-GCN}\\ w/G$^{3}$CN}& \makecell { \textcolor{red}{91.1}\\ $\uparrow$ \textbf{ \textcolor{red}{1.3}}} &  \makecell {91.1\\ $\uparrow$ 0.5}    &  \makecell { 92.2\\ $\uparrow$ 0.6}     &    \makecell { 92.9\\ $\uparrow$ 0.2}   &  \makecell {95.3 \\ $\uparrow$0.1}   &  \makecell {95.7  \\$\uparrow$0.2}    &  \makecell{ 96.5}     &    \makecell{ 96.9 }    &   \makecell { \textcolor{red}{87.4}\\ $\uparrow$ \textbf{ \textcolor{red}{2.3}} }&\makecell { \textcolor{red}{88.3}\\ $\uparrow$ \textbf{ \textcolor{red}{1.0 }} }  & \makecell { \textcolor{red}{89.4}\\ $\uparrow$ \textbf{ \textcolor{red}{0.9}} }      & \makecell { \textcolor{red}{90.2}\\ $\uparrow$ \textbf{ \textcolor{red}{0.8 } } }    &  \makecell { \textcolor{red}{88.6}\\ $\uparrow$ \textbf{ \textcolor{red}{2.3}} }    &  \makecell { \textcolor{red}{89.2}\\ $\uparrow$ \textbf{ \textcolor{red}{0.7}} }    & \makecell { \textcolor{red}{90.4}\\ $\uparrow$ \textbf{ \textcolor{red}{0.7}} }  &\makecell { \textcolor{red}{91.3}\\ $\uparrow$ \textbf{ \textcolor{red}{0.6 } } } \\
     
    \bottomrule
    \end{tabular}
    }
  \label{tab:addlabel}%
  \vspace{-1em}
\end{table*}%

\noindent

\begin{table}[t]
\centering
\captionof{table}{Top-1 accuracy (\%) of different skeleton-based action recognition methods on the NW-UCLA dataset. 
}

\scalebox{0.75}{
\begin{tabular}{c||c|c|c|c}
   
    \toprule
    \multicolumn{1}{c||}{\multirow{2}{*}{\textbf{Method}}} & \multicolumn{4}{c}{\textbf{NW-UCLA}}  \\     
     \cmidrule{2-5}   
     \multicolumn{1}{c||}{} & \multicolumn{1}{c|}{J} & \multicolumn{1}{c} 
      {B} & \multicolumn{1}{c|}{J+B} & \multicolumn{1}{c}{4S}  \\
    \midrule
      DC-GCN+ADG\cite{ChengADG}&-&-& 95.3&- \\
     Shift-GCN\cite{ShiftGCN} &92.5& -  &94.2 &94.6\\
      FR-Head \cite{ZhouFrhead}&-&  - &  -   & 96.8      \\
      SkeletonGCL\cite{gcl}&95.1& 95.0  & 96.1    & 96.8      \\
      SelfGCN\cite{Self-GCN}&-& - & -    & 96.8      \\
     
      \midrule
 2s-AGCN\cite{2SShicv24}&92.0& 92.2 &95.0 &95.5   \\
 \rowcolor{gray!40}\makecell { 2s-AGCN \cite{2SShicv24}  w/G$^{3}$CN}&  \makecell {93.8  ($\uparrow$  \textbf{1.8 })  } & \makecell { 93.1  ($\uparrow$\textbf{0.9 })   }  &  \makecell {95.2 ($\uparrow$\textbf{0.2 })  }    &   \makecell {95.7 ($\uparrow$\textbf{0.2}) }     \\
    \midrule
    CTR-GCN\cite{Chenctrgcn} &94.6&  91.8&  94.2  &  96.5  \\
    \rowcolor{gray!40}\makecell {CTR-GCN\cite{Chenctrgcn} w/G$^{3}$CN}& \makecell { 94.6   } & \makecell { 92.2 ($\uparrow$\textbf{0.4 })   }  &  \makecell {94.2 }    &   \makecell {96.5 }     \\
     \midrule
  InfoGCN\cite{Info-GCN} &94.0& 95.3& 96.3& 97.0 \\
    \rowcolor{gray!40} \makecell {InfoGCN\cite{Info-GCN}  w/G$^{3}$CN}&\makecell {94.2($\uparrow$\textbf{0.2})  } &  \makecell {95.5 ($\uparrow$\textbf{0.2 })  } & \makecell {  \textcolor{red}{97.2} ($\uparrow$\textbf{\textcolor{red}{0.9 }})   }  &  \makecell {\textcolor{red}{97.2} ($\uparrow$\textbf{\textcolor{red}{0.2} })  }     \\
     
    \bottomrule
    \end{tabular}
    }
    
   \label{tab:addlabel-1}
  \vspace{-1em}
\end{table}%

\begin{table}[t]
\centering
 \captionof{table}{Impact of Gaussian Topology and Gated GCN, along with three representative GCN-based backbones, on the NTU60 dataset under the X-Sub setting using the joint modality (J)(\%). 
}

\scalebox{0.65}{
 \begin{tabular}{c||c|c|c|c|c|c|c|c|c}
 
    \toprule
    \multicolumn{1}{c||}{\multirow{2}{*}{\textbf{Model}}} & \multicolumn{3}{c}{\textbf{2s-AGCN\cite{2SShicv24} }}& \multicolumn{3}{c}{\textbf{CTR-GCN}\cite{Chenctrgcn} }& \multicolumn{3}{c}{\textbf{InfoGCN\cite{Info-GCN} }}  \\  
    \cmidrule{2-10}   
     \multicolumn{1}{c||}{} & \multicolumn{1}{c|}{Acc.} & \multicolumn{1}{c|} 
      {Para.} & \multicolumn{1}{c|}{FLOPs} & \multicolumn{1}{c|}{Acc.} & \multicolumn{1}{c|} 
      {Para.} & \multicolumn{1}{c|}{FLOPs} & \multicolumn{1}{c|}{Acc.} & \multicolumn{1}{c|} 
      {Para.} & \multicolumn{1}{c}{FLOPs}   \\
      
      \midrule

Baseline&88.9&3.76M & 3.98G& 89.7&1.42M &1.78G &  89.8&1.54M &1.66G\\
\midrule
 \rowcolor{gray!40} \makecell {+Gaussian\\ Topology}&  \makecell { 90.1\\($\uparrow$\textbf{1.2 })   }& 4.80M& 5.26G&  \makecell {90.4\\($\uparrow$\textbf{0.7 })  }   &1.57M &1.96G &   \makecell {91.0\\($\uparrow$\textbf{1.2 }) }& 2.57M&2.94G\\
    \midrule
     \rowcolor{gray!40} +Gated GCN&  \makecell { 89.6\\($\uparrow$\textbf{0.7 })   }&9.39M &11.69G &  \makecell {90.5\\($\uparrow$\textbf{0.8})  }    &6.17M& 8.39G &   \makecell {91.1\\($\uparrow$\textbf{1.3 }) } &7.06M &9.13G\\
      \midrule
     \rowcolor{gray!40} \makecell { +Gaussian \\Topology\\+Gated GCN}&  \makecell { 90.1\\($\uparrow$\textbf{1.2 })}& 9.54M& 11.87G &  \makecell {90.8\\($\uparrow$\textbf{1.1 })  }&6.32M &8.56G &  \makecell {91.1\\($\uparrow$\textbf{1.3 }) } & 7.24M &9.31G\\
    \bottomrule
   
    \end{tabular}

  }
  \label{tab:addlabel-2}
  \vspace{-2em}
\end{table}%

\textbf{NTU RGB+D}. NTU RGB+D (NTU60) \cite{ntu60} consists of 56,880 samples, each representing a sequence of skeleton motion with 25 joints, across 60 action categories. The actions are performed by 40 subjects and captured by 3 Kinect cameras from different viewpoints. Two evaluation protocols are used: cross-subject (X-Sub), where training data comes from 20 subjects and testing data from another 20, and cross-view (X-View), where data from cameras 2 and 3 are used for training, and data from camera 1 for testing.

\textbf{NTU RGB+D 120}. NTU RGB+D 120 (NTU120) \cite{ntu120} is an expansion of NTU RGB+D, this dataset  adds 60 more action categories and 57,367 skeleton sequences. It includes 106 subjects and uses two protocols: cross-subject (X-Sub), with 53 subjects for training and 53 for testing, and cross-setup (X-Set), where data with even setup IDs are used for training, and odd setup IDs for testing.

\textbf{Northwestern-UCLA}. The Northwestern-UCLA (NW-UCLA) dataset \cite{ucla} contains 1,494 samples across 10 action categories, performed by 10 subjects and captured by 3 Kinect cameras. It uses data from the first two cameras for training and the third for testing, with skeletons annotated with 20 joints.

\subsection{Implementation Details}
To comprehensively validate G$^{3}$CN, we compare it against three  GCNs (CTR-GCN \cite{Chenctrgcn}, 2S-AGCN \cite{2SShicv24},  and InfoGCN \cite{Info-GCN}) as baseline models. Specifically, this involves replacing the corresponding GCN modules in these three baseline models with our proposed G$^{3}$CN, and then comparing the experimental results with the original ones. The implementations of experiments vary across these different models. 

\textbf{CTR-GCN and 2s-AGCN:}  For CTR-GCN and 2s-AGCN baseline models, when replacing the GCN modules in these models with our proposed G$^{3}$CN, all implementations follow their original training recipes, except for some implementations as follows: 1) The training epoch is set to 85; 2)The initial learning rate is 0.05 and decays by a factor of 0.1 at epochs 45, 65, and 75.   

\textbf{InforGCN:}For InfoGCN, when replacing the GCN modules in these models with our proposed G$^{3}$CN, all implementations follow their training recipes, except for the learning rate, which is set to 0.05 in this paper.

\textbf{Feature dimension reduction factor:}  Linear transformations $\psi_1$, $\psi_2$, $\psi_3$, and $\psi_4$ are used to reduce feature dimensionality. Following the implementation details in CTR-GCN, the corresponding reduction factor $r = C / C^{\prime\prime}$ in G$^{3}$CN is set to 8 in all layers except the input layer for all experiments.

All experiments are conducted on two RTX 3090 GPUs using the PyTorch deep learning framework.

\begin{table*}[t]
\centering
 \caption{Recognition accuracy on Ambiguous Actions .  The experiment is conducted on NTU60 under the X-Sub setting using the joint modality (J)
}
    
    \scalebox{0.83}{
\begin{tabular}{c||c|c|c|c|c}  
 \toprule
Model & Class &  Acc& \multicolumn{3}{c}{ Wrong Class: ratio} \\
\midrule
\multirow{6}{*}[-5.5ex]{CTR-GCN\cite{Chenctrgcn}} & writing & 42.4\%& \makecell{typing on a keyboard: 41.82\%} & reading: 26.36\% & \makecell{ playing with phone/tablet: 24.55\%} \\
\cmidrule{2-6}
 & reading & 50.0\%& writing: 37.36\% &  \makecell{playing with phone/tablet: 30.77\%} &  \makecell{typing on a keyboard: 9.89\%} \\
 \cmidrule{2-6}
 & \makecell{playing with phone/tablet }& 54.0\%& reading: 30.49\% & \makecell{ typing on a keyboard: 25.61\%} & writing: 18.29\% \\
 \cmidrule{2-6}
 &\makecell{ typing on a keyboard} & 55.8\%& writing: 47.44\% & reading: 12.82\% & \makecell{playing with phone/tablet: 11.54\%} \\
 \cmidrule{2-6}
 & wear a shoe & 64.4\%& take off a shoe: 89.83\% &  \textbf{---}&  \textbf{---}\\
 \cmidrule{2-6}
 & take off a shoe & 66.0\%& wear a shoe: 60.71\% &  \textbf{---}&  \textbf{---}\\
\midrule
\multirow{6}{*}[-5.5ex]{\makecell {\textbf{CTR-GCN\cite{Chenctrgcn}}\\textbf{w/}\textbf{G$^{3}$CN}}} & writing & 43.9\%($\uparrow$\textbf{1.5 })&  \makecell{typing on a keyboard: 42.45\%} & reading: 26.42\% & \makecell{ playing with phone/tablet: 22.64\%} \\
\cmidrule{2-6}
 & reading & 53.3\%($\uparrow$\textbf{3.3 })& writing: 36.14\% & \makecell{ playing with phone/tablet: 32.53\%} &  \makecell{typing on a keyboard: 12.05\%} \\
 \cmidrule{2-6}
 & \makecell{playing with phone/tablet}& 63.20\%($\uparrow$\textbf{\textcolor{red}{8.2} })& writing: 29.03\% & reading: 22.58\% & \makecell{ typing on a keyboard: 16.13\%} \\
 \cmidrule{2-6}
 & \makecell{typing on a keyboard }& 58.9\%($\uparrow$\textbf{3.1 })& writing: 39.44\% & reading: 16.90\% & \makecell{ playing with phone/tablet: 14.08\%} \\
 \cmidrule{2-6}
 & wear a shoe & 74.4\%($\uparrow$\textbf{\textcolor{red}{10.0} })& take off a shoe: 80.00\% &  \textbf{---}&  \textbf{---}\\
 \cmidrule{2-6}
 & take off a shoe & 68.6\%($\uparrow$\textbf{2.8 })& wear a shoe: 70.59\% &  \textbf{---}&  \textbf{---}\\

 \bottomrule
\end{tabular}
}
 \label{tab:addlabel-3}%
 \vspace {-5mm}
\end{table*}

\begin{figure*}[b]
    \centering
    \includegraphics[width=0.80\linewidth]{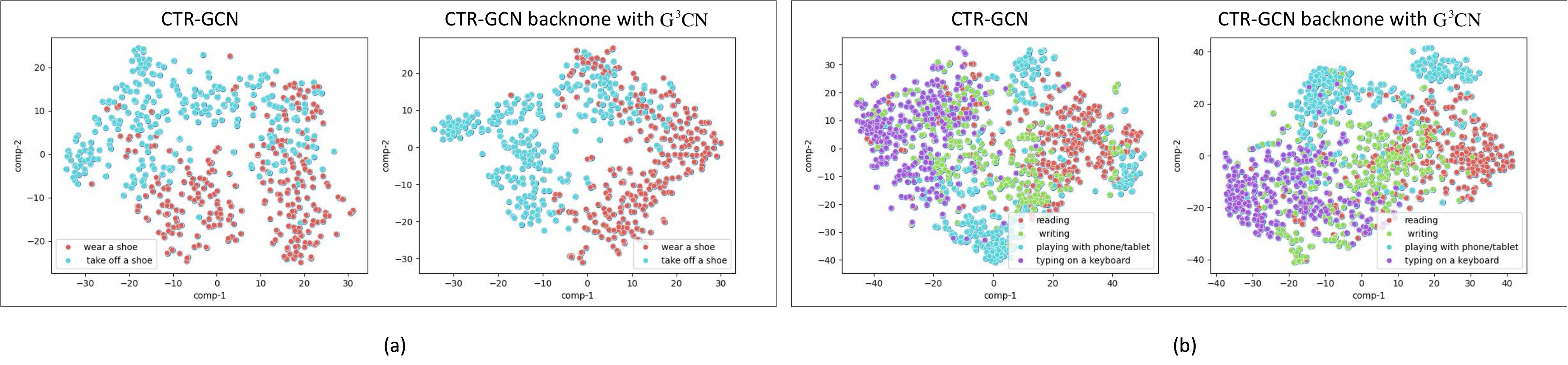}
 \vspace {-0.5em}   
    \caption{Visualization of latent representation by t-SNE for ambiguous
groups from the NTU60 dataset. Figures in (a) show visualization results for the group B (`wear a shoe' and `take off a shoe'), and figures in (b) show the visualization results for the group A (`writing', `reading', `playing with phone/tablet' and `typing on a keyboard'). The images on the left in (a) and (b) are from the CTR-GCN \cite{Chenctrgcn}, while the images on the right are from our method.}
    \label{fig:enter-label-sne}
 \vspace {-1.0em}
\end{figure*}

\subsection{Performance Comparison}
We integrate our method into three GCN-based models and compare them with state-of-the-art (SoTA) GCN methods by replacing the corresponding GCN modules with our G$^{3}$CN. A comparative analysis is provided in Tables \ref{tab:addlabel} and \ref{tab:addlabel-1}. We evaluate using four modalities: joint stream' (J), bone stream' (B), joint motion', and bone motion'. J + B represents the ensemble of J and B, and the 4-stream ensemble, denoted as 4S, includes all four modalities.

\textbf{NTU60 and NTU120.} Table \ref{tab:addlabel} lists the results on NTU60 and NTU120. From the results, we find that: 
\begin{itemize}
\item [(1)]  When replaced with G$^{3}$CN, all three baseline models exhibit notable enhancements across these two benchmarks across different  modalities and settings. For instance, when considering the J modality on NTU60 XSub, 2S-AGCN improves by 1.2\% (from 88.9\% to 90.1\%), CTR-GCN by 1.1\% (from 89.7\% to 90.8\%), and InfoGCN by 1.3\% (from 89.8\% to 91.1\%). Given that NTU60 is a well-established benchmark dataset, achieving such improvements is particularly challenging.

\item [(2)]Replaced with G$^{3}$CN,  InfoGCN improves by  2.3\% (from 85.1\% to 87.4\%), 1.0\% (from 87.3\% to 88.3\%), 0.9\% (from 88.5\% to 89.4\%), 0.8\% (from 89.4\% to 90.2\%), 2.3\% (from 86.3\% to 88.6\%), 0.7\% (from 88.5\% to 89.2\%), 0.7\% (from 89.7\% to 90.4\%) and 0.6\% (from 90.7\% to 91.3\%) respectively on the NTU120 dataset under X-Sub and X-Set settings using all 4 protocols ( J, B, J + B, and 4S).
\end{itemize}

\textbf{NW-UCLA.} Table \ref{tab:addlabel-1} displays the results for NW-UCLA. G$^{3}$CN consistently shows improvements across all three models. In GCN-based methods, by replacing the corresponding GCN  with G$^{3}$CN, InfoGCN achieves state-of-the-art performance on the NW-UCLA dataset using both modalities (J + B and 4S).

\begin{figure*}[b]
    \centering
    \includegraphics[width=0.80\linewidth]{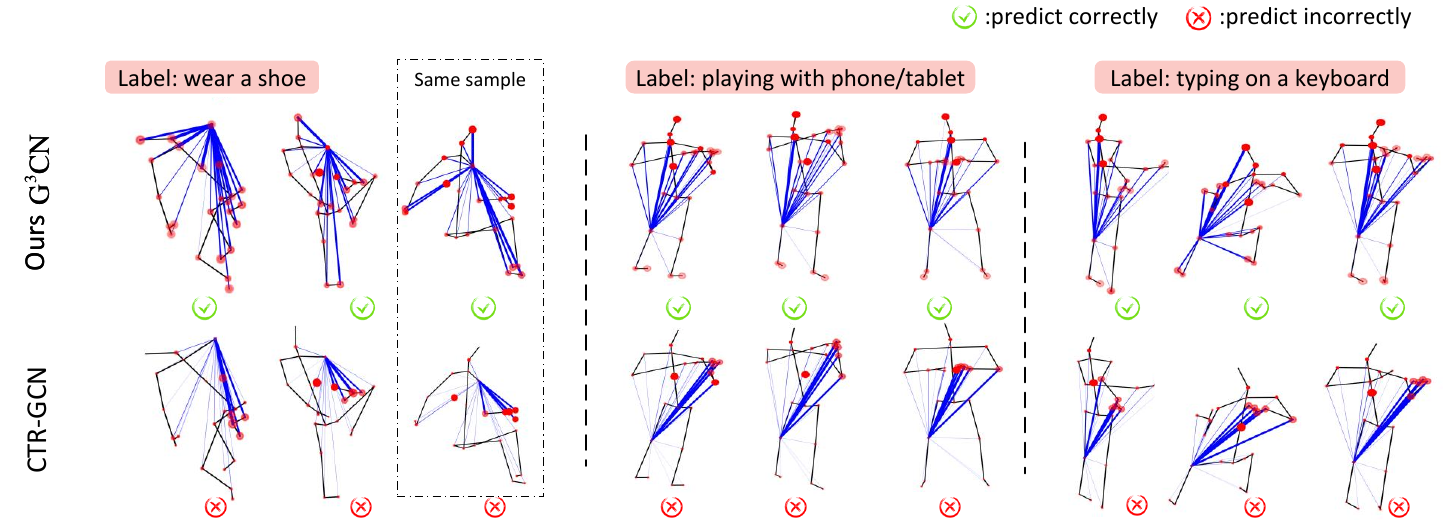}
 \vspace {-0.5em}   
    \caption{Topology visualization of samples from three ambiguous actions shows correlations between specific joints. For "playing with phone/tablet" and "typing on a keyboard," correlations with the "left knee" are visualized, while for "wear a shoe," correlations with the "left shoulder" are shown. The size of the red circles and width of the blue lines represent the strength of these correlations.}
    \label{fig:enter-label-sample}
 \vspace {-5mm}
\end{figure*}

\subsection{Ablation Study}
In this section, we perform diagnostic experiments to verify the design of G$^{3}$CN. Specifically, we utilize 2S-AGCN \cite{2SShicv24}, CTR-GCN \cite{Chenctrgcn}, and InfoGCN \cite{Info-GCN} as the backbone models for experiments on the NTU60 dataset under the X-Sub setting, focusing on the joint modality (J). In Table \ref{tab:addlabel-2}, we evaluate the effectiveness of Gaussian Topology and Gated GCN, reporting both model performance and computational metrics (parameters and FLOPs). We observe that both components enhance recognition across all three backbones, and their combined use significantly boosts performance, highlighting their complementary nature.

\textbf{ Gaussian Topology.}  When Gaussian Topology is introduced, 2S-AGCN improves by 1.2\% (from 88.9\% to 90.1\%), CTR-GCN by 0.7\% (from 89.7\% to 90.4\%), and InfoGCN by 1.2\% (from 89.8\% to 91.0\%).

\textbf{ Gated GCN.}  Introducing Gated GCN results in improvements of 0.7\% for 2S-AGCN (from 88.9\% to 89.6\%), 0.8\% for CTR-GCN (from 89.7\% to 90.5\%), and 1.3\% for InfoGCN (from 89.8\% to 91.1\%).

\textbf{ Gaussian Topology+Gated GCN.}  Using both Gaussian Topology and Gated GCN, 2S-AGCN is improved by 1.2\% (from 88.9\% to 90.1\%), CTR-GCN by 1.1\% (from 89.7\% to 90.8 \%), and InfoGCN by 1.3\% (from 89.8\% to 91.1\%).

\subsection{Performance on Ambiguous Actions}
In our analysis of the NTU60 dataset using CTR-GCN, we identified two ambiguous action groups: (A) writing, reading, playing with phone/tablet, and typing on a keyboard, and (B) wear a shoe and take off a shoe. These actions were prone to misclassification. For instance, "playing with phone/tablet" in group A was often misclassified as "writing", "reading", or "typing on a keyboard". The recognition results for these ambiguous groups are displayed in  table \ref{tab:addlabel-3}.  The experiment is conducted on NTU60 under the X-Sub setting using the joint modality (J).

The experiment results show that applying our method led to improvements in recognition accuracy for all actions. In group A, with "playing with phone/tablet" improving by 8.2\% (from 54.0\% to 63.2\%), "typing on a keyboard" improving by 3.1\% (from 55.8\% to 58.9\%),`reading' improves by 3.3\% (from 50.0\% to 53.3\%), and `writng' improves by 1.5\% (from 42.4\% to 43.9\%). In group B, the recognition accuracy for "wear a shoe" increased by 10.0\% (from 64.4\% to 74.4\%).

After applying G$^{3}$CN, the distribution of misclassifications changes with improved recognition accuracy. For example, the probability of misclassifying "reading" as "writing" decreases, but it becomes more likely to be misclassified as "playing with phone/tablet" or "typing on a keyboard" (both involving simpler hand movements). This suggests that our method effectively distinguishes the topological relationships between hand joints and other joints in "writing" and other ambiguous actions, improving recognition accuracy.

We compare our approach with CTR-GCN \cite{Chenctrgcn} by visualizing the distribution of ambiguous groups in the feature space using t-SNE. As shown in Fig.\ref{fig:enter-label-sne}, our model achieves a more discriminative representation with compact clustering.


Additionally, we also visualize the topology graph generated by the G$^{3}$CN model by averaging the graphs from the final basic block: $\bar{\bm{A}}_{Gaussian} = \frac{1}{M}\sum_{m=1}^{M}\bm{A}_{Gaussian}^{(m)}$, where $M$ is the number of G$^{3}$CNs in the block. A similar process is applied to the CTR-GCN model. The results in Fig.\ref{fig:enter-label-sample} show, for each action, topology graphs generated by both G$^{3}$CN and CTR-GCN for the same sample.

From the visualizations, we observe that: (a) G$^{3}$CN correctly identifies some samples that CTR-GCN fails to recognize, and (b) G$^{3}$CN enhances weak joint correlations learned by CTR-GCN, improving recognition accuracy. Specifically:
\begin{itemize}
\item [(1)] \textbf{`wear a shoe'.} For samples belonging to this class, the movement of the feet is crucial for sample identification. However, the correlations between the joints on both feet and `left shoulder' learned by CTR-GCN \cite{Chenctrgcn} is very weak. Our proposed algorithm corrects these correlations, leading to improved recognition accuracy for the `wear a shoe' action category.
As shown in Fig.\ref{fig:enter-label-sample}, some samples that CTR-GCN cannot correctly identify achieve accurate recognition results using our proposed approach G$^{3}$CN;
\item [(2)] \textbf{`playing with phone/tablet' and `typing on a keyboard'.} For samples belonging to these classes, movement of both hands is crucial for sample identification. However, visualization results of topology graphs learned by CTR-GCN \cite{Chenctrgcn} show strong correlations between joints of one hand and the `left knee', while correlations between joints of the other hand and the `left knee' are weak.  G$^{3}$CN corrects these weak correlations, improving recognition for these actions, as depicted in Fig.\ref{fig:enter-label-sample}.
\end{itemize}

\section{CONCLUSIONS}
In this paper, we present a novel approach called Gaussian Topology Refinement Gated Graph Convolution (G$^{3}$CN) for skeleton-based action recognition.  G$^{3}$CN   consists of two main components:  Gaussian Topology and Gated Graph Convolution. The Gaussian Topology module learns refined topologies by using a Gaussian filter to process the topology graph. Meanwhile, the Gated Graph Convolution module ensures that the aggregated features are sparser compared to traditional GCNs by introducing gating mechanisms. Extensive experiments on three widely used benchmarks demonstrate that our G$^{3}$CN can effectively enhance the representation ability of GCN networks for mbiguous actions. Moreover, G$^{3}$CN is applicable to most GCN-based backbones, highlighting its versatility and effectiveness.


\begin{thebibliography}{99}


\bibitem{Chenctrgcn}
Y.~Chen, Z.~Zhang, C.~Yuan, B.~Li, Y.~Deng, and W.~Hu, ``Channel-wise topology refinement graph convolution for skeleton-based action recognition,'' \emph{In Proceedings of the IEEE/CVF International Conference on Computer Vision}, pp. 13\,359--13\,368, 2021.

\bibitem{MST-GCN}
Z.~Chen, S.~Li, B.~Yang, Q.~Li, and H.~Liu, ``Multi-scale spatial temporal graph convolutional network for skeleton-based action recognition,'' \emph{In Proceedings of the AAAI Conference on Artificial Intelligence}, vol.~35, pp. 1113--1122, 2021.

\bibitem{lstm}
J.~Liu, A.~Shahroudy, D.~Xu, A.~C. Kot, and GangWang, ``Skeleton-based action recognition using spatiotemporal lstm network with trust gates,'' \emph{IEEE transactions on pattern analysis and machine intelligence}, vol.~40, no.~12, pp. 3007--3021, 2017.

\bibitem{hh-to}
L.~Ke, K.-C. Peng, and S.~Lyu, ``Towards to-at spatio-temporal focus for skeleton-based action recognition,'' \emph{arXiv preprint arXiv:2202.02314}, 2022.


\bibitem{gcl-sk1}
H.~Duan, Y.~Zhao, K.~Chen, D.~Lin, and B.~Dai, ``Revisiting skeleton-based action recognition,'' \emph{In Proceedings of the IEEE/CVF Conference on Computer Vision and Pattern Recognition}, pp. 2969--2978, 2022.

\bibitem{Liucv20}
Z.~Liu, H.~Zhang, Z.~Chen, Z.~Wang, and W.~Ouyang, ``Disentangling and unifying graph convolutions for skeleton-based action recognition,'' \emph{In Proceedings of the IEEE/CVF Conference on Computer Vision and Pattern Recognition}, pp. 143--152, 2020.



\bibitem{Ddgcn}
M.~Korban and X.~Li, ``Ddgcn: A dynamic directed graph convolutional network for action recognition,'' \emph{In European Conference on Computer Vision}, pp. 761--776, 2020.

\bibitem{Zhaocv36}
R.~Zhao, KangWang, H.~Su, and Q.~Ji, ``Bayesian graph convolution lstm for skeleton based action recognition,'' \emph{In Proceedings of the IEEE/CVF International Conference on Computer Vision}, pp. 6882--6892, 2019.

\bibitem{Tangcv27}
Y.~Tang, Y.~Tian, J.~Lu, P.~Li, and J.~Zhou, ``Deep progressive reinforcement learning for skeleton-based action recognition,'' \emph{In Proceedings of the IEEE Conference on Computer Vision and Pattern Recognition}, pp. 5323--5332, 2018.

\bibitem{2SShicv24}
L.~Shi, Y.~Zhang, J.~Cheng, and H.~Lu, ``Two stream adaptive graph convolutional networks for skeleton based action recognition,'' \emph{In Proceedings of the IEEE Conference on Computer Vision and Pattern Recognition}, pp. 12\,026--12\,035, 2019.

\bibitem{Info-GCN}
H.~gun Chi, M.~H. Ha, S.~Chi, S.~W. Lee, Q.~Huang, and K.~Ramani, ``Infogcn: Representation learning for human skeleton-based action recognition,'' \emph{In Proceedings of the IEEE/CVF Conference on Computer Vision and Pattern Recognition}, pp. 20\,186--20\,196, 2022.

\bibitem{Cho}
K.~Cho, B.~van Merrienboer, C.~Gulcehre, D.~Bahdanau, F.~Bougares, H.~Schwenk, and Y.~Bengio, ``Learning phrase representations using rnn encoder-decoder for statistical machine translation,'' \emph{arXiv preprint arXiv:1406.1078}, 2014.

\bibitem{Yancv32}
S.~Yan, Y.~Xiong, and D.~Lin, ``Spatial temporal graph convolutional networks for skeleton-based action recognition,'' \emph{In Proceedings of the AAAI Conference on Artificial Intelligence}, pp. 7444--7452, 2018.




\bibitem{Zhangpr42}
X.~Zhang, C.~Xu, X.~Tian, and D.~Tao, ``Graph edge convolutional neural networks for skeletonbased action recognition,'' \emph{IEEE Transactions on Neural Networks and Learning Systems}, vol.~31, no.~8, pp. 3047--3060, 2019.

\bibitem{Huangcv9}
Z.~Huang, X.~Shen, X.~Tian, H.~Li, J.~Huang, and X.-S. Hua, ``Spatio-temporal inception graph convolutional networks for skeleton-based action recognition,'' \emph{In Proceedings of the 28th ACM International Conference on Multimedia}, pp. 2122--2130, 2020.

\bibitem{Licv15}
M.~Li, S.~Chen, X.~Chen, Y.~Zhang, Y.~Wang, and Q.~Tian, ``Actional-structural graph convolutional networks for skeleton-based action recognition,'' \emph{In Proceedings of the IEEE/CVF Conference on Computer Vision and Pattern Recognition}, pp. 3595--3603, 2019.

\bibitem{Zhangiccv35}
P.~Zhang, C.~Lan, W.~Zeng, J.~Xing, J.~Xue, and N.~Zheng, ``Semantics-guided neural networks for efficient skeleton-based human action recognition,'' \emph{In Proceedings of the IEEE/CVF Conference on Computer Vision and Pattern Recognition}, pp. 1112--1121, 2020.

\bibitem{Dynamic-GCN}
F.~Ye, S.~Pu, Q.~Zhong, C.~Li, D.~Xie, , and H.~Tang, ``Dynamic gcn: Context-enriched topology learning for skeleton-based action recognition,'' \emph{In Proceedings of the 28th ACM International Conference on Multimedia}, pp. 55--63, 2020.

\bibitem{ChengADG}
K.~Cheng, Y.~Zhang, C.~Cao, L.~Shi, J.~Cheng, and H.~Lu, ``Decoupling gcn with dropgraph module for skeleton-based action recognition,'' \emph{In European Conference on Computer Vision}, pp. 536--553, 2020.

\bibitem{PangIGformer}
Y.~Pang, Q.~Ke, H.~Rahmani, J.~Bailey, and J.~Liu, ``Igformer: Interaction graph transformer for skeleton-based human interaction recognition,'' \emph{In Proceedings of the European Conference on Computer Vision (ECCV)}, pp. 605--622, 2022.

\bibitem{gcl-contrastive}
T.~Guo, H.~Liu, Z.~Chen, M.~Liu, T.~Wang, and R.~Ding, ``Contrastive learning from extremely augmented skeleton sequences for self-supervised action recognition,'' \emph{In Proceedings of the AAAI Conference on Artificial Intelligence}, pp. 762--770, 2022.

\bibitem{ZhouFrhead}
H.~Zhou, Q.~Liu, , and Y.~Wang, ``Learning discriminative representations for skeleton based action recognition,'' \emph{In Proceedings of the IEEE Conference on Computer Vision and Pattern Recognition (CVPR)}, pp. 10\,608--10\,617, 2023.

\bibitem{gcl}
X.~Huang, H.~Zhou, B.~Feng, X.~Wang, W.~Liu, J.~Wang, H.~Feng, J.~Han, E.~Ding, and J.~Wang, ``Graph contrastive learning for skeleton-based action recognition,'' \emph{The International Conference on Learning Representations (ICLR)}, 2023.

\bibitem{Henaff}
M.~Henaff, J.~Bruna, , and Y.~LeCun, ``Deep convolutional networks on graphstructured data,'' \emph{In arXiv:1506.05163}, 2015.

\bibitem{Duvenaud}
D.~Duvenaud, D.~Maclaurin, J.~Aguilera-Iparraguirre, R.~Gómez-Bombarelli, T.~Hirzel, A.~Aspuru-Guzik, , and R.~P. Adams, ``Convolutional networks on graphs for learning molecular fingerprints,'' \emph{Advances in Neural Information Processing Systems (NIPS)}, 2015.


\bibitem{Kipf}
T.~N. Kipf and M.~Welling, ``Semi-supervised classification with graph convolutional networks,'' \emph{The International Conference on Learning Representations (ICLR)}, 2017.

\bibitem{Stigler1982}
S.~M. Stigler, ``A modest proposal: a new standard for the normal,'' \emph{The American Statistician}, vol.~36, no.~2, pp. 137--138, 1982.

\bibitem{ntu60}
A.~Shahroudy, J.~Liu, T.-T. Ng, and G.~Wang, ``Ntu rgb+ d: A large scale dataset for 3d human activity analysis,'' \emph{In Proceedings of the IEEE Conference on Computer Vision and Pattern Recognition}, pp. 1010--1019, 2016.

\bibitem{ntu120}
A.~Shahroudy, M.~Perez, G.~Wang, L.-Y. Duan, and A.~C. Kot, ``Ntu rgb+ d 120: A large-scale benchmark for 3d human activity understanding,'' \emph{IEEE Transactions on Pattern Analysis and Machine Intelligence}, vol.~42, no.~10, pp. 2684--2701, 2019.

\bibitem{ucla}
J.~Wang, X.~Nie, Y.~Xia, Y.~Wu, and S.-C. Zhu, ``Cross-view action modeling, learning and recognition,'' \emph{In Proceedings of the IEEE Conference on Computer Vision and Pattern Recognition}, pp. 2649--2656, 2014.

\bibitem{HeCV8}
K.~He, X.~Zhang, S.~Ren, and J.~Sun, ``Deep residual learning for image recognition,'' \emph{In Proceedings of the IEEE/CVF Conference on Computer Vision and Pattern Recognition}, pp. 770--778, 2016.

\bibitem{ShiftGCN}
K.~Cheng, Y.~Zhang, X.~He, W.~Chen, J.~Cheng, and H.~Lu, ``Skeleton-based action recognition with shift graph convolutional network,'' \emph{In Proceedings of the IEEE/CVF Conference on Computer Vision and Pattern Recognition}, pp. 183--192, 2020.


\bibitem{STCnet_2023_ICCV}
J.~Lee, M.~Lee, S.~Cho, S.~Woo, S.~Jang, and S.~Lee, ``Leveraging spatio-temporal dependency for skeleton-based action recognition,'' \emph{Proceedings of the IEEE/CVF International Conference on Computer Vision (ICCV)}, pp. 10\,255--10\,264, 2023.

\bibitem{HDGCN_2023_ICCV}
J.~Lee, M.~Lee, D.~Lee, and S.~Lee, ``Hierarchically decomposed graph convolutional networks for skeleton-based action recognition,'' \emph{Proceedings of the IEEE/CVF International Conference on Computer Vision (ICCV)}, pp. 10\,444--10\,453, 2023.

\bibitem{Self-GCN}
Z.~Wu, P.~Sun, X.~Chen, K.~Tang, T.~Xu, L.~Zou, X.~Wang, M.~Tan, F.~Cheng, T.~Weise, ``SelfGCN: Graph Convolution Network With Self-Attention for Skeleton-Based Action Recognition,'' \emph{IEEE Transactions on Image Processing},vol.~33, pp. 4391--4403, 2024.

\bibitem{DeGCN}
 W.~Myung, N.~Su, J.~Xue, G.~ Wang, ``DeGCN: Deformable Graph Convolutional
 Networks for Skeleton-Based Action Recognition,'' \emph{IEEE Transactions on Image Processing},vol.~33, pp. 2477--2490, 2024.

\bibitem{Large-kernel}
 Y.~Liu, H.~Zhang, Y.~Li, K.~ He, D.~ Xu, ``Skeleton-based Human Action Recognition via Large-kernel Attention Graph Convolutional Network,'' \emph{IEEE Transactions on Visualization and Computer Graphics},vol.~29, no.~5, pp. 2575--2585, 2023.




\end{thebibliography}
\end{document}